\documentclass[10pt]{article} % For LaTeX2e
\usepackage[accepted]{tmlr}
\usepackage{amsmath,amsfonts,bm}

\def\eqref#1{equation~\ref{#1}}
\def\1{\bm{1}}

\DeclareMathAlphabet{\mathsfit}{\encodingdefault}{\sfdefault}{m}{sl}
\SetMathAlphabet{\mathsfit}{bold}{\encodingdefault}{\sfdefault}{bx}{n}

\usepackage{hyperref}
\usepackage{url}
\usepackage{subfig}
\usepackage{booktabs}
\usepackage{graphicx}
\usepackage{caption}
\usepackage{wrapfig}
\usepackage{bm}
\usepackage{longtable}
\usepackage{enumitem}
\usepackage{amsmath}
\usepackage{color}
\usepackage{xcolor}

\newcommand{\valKodakCoinFirstLayerIrrErrEV}{66.0}
\newcommand{\valKodakCoinImageSpecializedPEAdvantage}{0.23}

\newcommand{\valKodakCoinIrrErr}{0.18}
\newcommand{\valKodakCoinIrrErrWithFixedPE}{0.10}

\newcommand{\valKodakCoinTransferredPEAdvantage}{0.14}

\newcommand{\valLastThousandStepsMedianImprovement}{0.0016}
\newcommand{\valLastThousandStepsTailImprovement}{0.070}

\newcommand{\valSingleArchMean}{30.53}

\newcommand{\valSingleArchStd}{4.41}

\newcommand{\valMultiArchMean}{28.12}
\newcommand{\valMultiArchStd}{4.86}

\newcommand{\etal}{et al.}
\title{Predicting the Encoding Error of SIRENs}

\author{\name Jeremy Vonderfecht \email vonder2@pdx.edu \\
      \addr Department of Computer Science \\ Portland State University
      \AND
      \name Feng Liu \email fliu@pdx.edu\\
      \addr Department of Computer Science \\ Portland State University}

\def\month{05}  % Insert correct month for camera-ready version
\def\year{2024} % Insert correct year for camera-ready version
\def\openreview{\url{https://openreview.net/forum?id=iKPC7N85Pf}} % Insert correct link to OpenReview for camera-ready version

\begin{document}

\maketitle

\begin{abstract}
Implicit Neural Representations (INRs), which encode signals such as images, videos, and 3D shapes in the weights of neural networks, are becoming increasingly popular. Among their many applications is signal compression, for which there is great interest in achieving the highest possible fidelity to the original signal subject to constraints such as neural network size, training (encoding) and inference (decoding) time. But training INRs can be a computationally expensive process, making it challenging to determine the best possible tradeoff under such constraints. Towards this goal, we present a method which predicts the encoding error that a popular INR network (SIREN) will reach, given its network hyperparameters and the signal to encode. This method is trained on a unique dataset of 300,000 SIRENs, trained across a variety of images and hyperparameters.\footnote{Dataset available here: \href{https://huggingface.co/datasets/predict-SIREN-PSNR/COIN-collection}{huggingface.co/datasets/predict-SIREN-PSNR/COIN-collection}} Our predictive method demonstrates the feasibility of this regression problem, and allows users to anticipate the encoding error that a SIREN network will reach in milliseconds instead of minutes or longer. We also provide insights into the behavior of SIREN networks, such as why narrow SIRENs can have very high random variation in encoding error, and how the performance of SIRENs relates to JPEG compression.
\end{abstract}

\section{Introduction}
\label{sec:intro}

Since the introduction of neural radiance fields in 2020, there has been a swell of interest in neural networks as representations of data. Such networks, commonly called \textit{implicit neural representations} (INRs), or \textit{coordinate networks}, are trained to fit relatively low-dimensional signals, such as 2D images, 3D shapes (typically as occupancy functions or signed distance functions), or neural radiance fields (NeRFs). Implicit neural representations have proven useful for a diverse range of tasks, including super-resolution \citep{chen2021learning}, novel view synthesis \citep{park2021nerfies}, video manipulation \citep{pumarola2021d, gao2021dynamic, mai2022motion, Li_2021_CVPR, xian2021space, tretschk2021non}, and compression.

There is great interest in achieving the highest possible fidelity to the encoded signal subject to constraints such as network size, training and inference time. In the context of signal compression, researchers wish to minimize the network's size while maintaining a low encoding error.

In this paper, we study SIREN networks trained to fit natural images. Multilayer perceptrons with periodic activation functions, or SIRENs, have many advantages. They are simple to implement and quick to train. They train stably even when 10+ layers deep, and have well-behaved derivatives \citep{sitzmann2020implicit}. Therefore, they have become a baseline architecture for INRs, for example, see \cite{dupont2021coin, martel2021acorn,lindell2022bacon, chen2021nerv}. Meanwhile, natural images have become a standard training target for the evaluation of coordinate network architectures, as seen in papers such as \cite{sitzmann2020implicit}, \cite{tancik2020fourier}, and \cite{dupont2021coin}. 

In the context of images, by far the most common metric by which SIRENs are judged is mean squared error (MSE), often formulated as peak signal-to-noise ratio (PSNR). Practitioners often seek to maximise their INRs' PSNR subject to constraints such as training time (for signal compression, this is encoding time), number of trainable parameters (for compression), or forward-pass computation time (for decoding efficiency). However, training SIRENs can be time-consuming. For example, training a SIREN to compress a 512x768 pixel image to 0.3 bits per pixel with the method from \cite{dupont2021coin} takes 50 minutes on a Titan X GPU. Suppose that we wish to compress an image using the smallest SIREN that will achieve a PSNR of at least 30 dB. Without a predictive model of SIREN performance, we will have to perform a time-consuming hyperparameter sweep to determine the best SIREN architecture for our task. A predictive model of what encoding error a given SIREN will reach will allow us to make quick, informed choices across a broad range of applications.

To address this issue, we develop a predictive model of how well a SIREN will encode a given image. Our model takes in a set of SIREN hyperparameters and the target image, and predicts the encoding error, measured as PSNR, that the SIREN will achieve. Within the training domain, our model makes accurate predictions of final training loss in just milliseconds, as opposed to minutes of training.

To train this model, we develop a dataset of 300,000 SIRENs, trained across 100,000 photographs and a range of SIREN hyperparameters. This dataset contains the hyperparameters, training loss curves, and final weights of each SIREN. Initial weights for each SIREN can be reconstructed using our source code. We have made this dataset publicly available.

Like all neural networks, our predictive model of SIREN encoding error is somewhat of a black box. Therefore, we also present a collection of experiments which offer more human-friendly insights into the relationship between the SIREN hyperparameters, the image to encode, and the SIREN's encoding error.

Our contributions are as follows:

\begin{itemize}
    \item A novel dataset of 300,000 SIRENs trained to encode RGB images. (Section \ref{sec:datasets})
    \item A neural-network-based predictive model of SIREN encoding error. (Section \ref{sec:predictive_models})
    \item A series of studies on SIRENs, which show that:
        \begin{itemize}
            \item Deep, narrow SIRENs like those used by \cite{dupont2021coin} have significant random variation in their encoding error, largely explained by the initialization of the first layer. (Section \ref{sec:random_variation})
            \item SIRENs follow scaling power-laws as described in \cite{bahri2021explaining}. (Section \ref{sec:power-laws})
            \item There is a strong correlation between JPEG and SIREN compression. (Section \ref{sec:jpeg})
            \item Unlike some other neural networks, SIREN training dynamics are \textbf{not} well-approximated by their empirical neural tangent kernels. (Section \ref{sec:extrapolating-ntk})
            \item low-bitrate SIRENs visibly compress the color space of the images they encode to a more restricted color palette. (Appendix \ref{sec:color-space})
        \end{itemize}
\end{itemize}

\begin{figure}[t]
    \centering
    \includegraphics[width=0.75\textwidth]{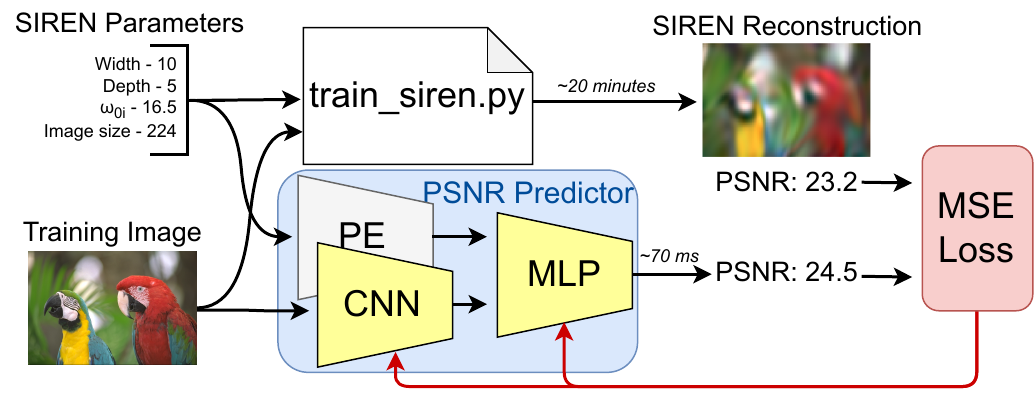}
    \caption{Overview of our many-architecture encoding error predictor. First, we train SIRENs on many images with a random sampling of hyperparameters. Then, our many-architecture predictor takes the same inputs as the SIREN training script, and predicts the SIRENs' PSNR. The training image is passed through a convolutional-network-based feature extractor (CNN), while the SIREN hyperparameters are fed through a positional encoding (PE). Then both are concatenated and passed through a fully-connected network (MLP) which predicts what PSNR the training script will reach.}
    \label{fig:overview}
\end{figure}

\section{Methods}

\subsection{A brief overview of the SIREN Architecture}
\label{sec:siren}

SIRENs, introduced by \cite{sitzmann2020implicit}, are simply multilayer perceptrons with sinusoidal activations. A SIREN \(\Phi_{\theta}(x)\) has the following form: 

\begin{align}
\begin{split}
    \Phi_{\theta}(x) &= W_L h_{L-1}(x) + b_L \\
    h_l(x) &= \sin(W_l h_{l-1}(x) + b_l)
    \quad \text{for } l = 1, \dots, L-1\\
    h_0(x) &= \omega_{0} x
\end{split}
\end{align}

where \( L \) is the number of layers, $\theta = (W_0, b_0, W_1, b_1, \dots, W_{L}, b_{L})$ are the network's trainable weight matrices and bias vectors, and \( \omega_0 \) is a scalar hyperparameter which controls the spatial frequency of activations in the first hidden layer \( h_1(x) \), and hence the spectral bias of the network.

SIRENs are commonly used as compressive representations of images. In this setup, a SIREN learns to represent a single image by learning a function from (x, y) pixel coordinates to (r, g, b) colors. Inputs and outputs are both normalized to [-1, 1]. The SIREN is trained with $L_2$ loss between predicted and true colors. By storing the learned weights at low precision, we can use the network as a compressed representation of the image. The encoding error of this representation, as measured in PSNR, is equivalent to the best $L_2$ loss achieved during training.

\subsection{Datasets}
\label{sec:datasets}

In this paper, we make use of two image datasets: \cite{kodak1991} and MSCOCO \citep{lin2014microsoft}. The Kodak dataset is a popular reference dataset in image compression literature, and we use it for our small-scale exploratory studies in Sections \ref{sec:random_variation}, \ref{sec:power-laws}, and \ref{sec:jpeg}. But the Kodak dataset consists of only 24 images, and our predictive models of SIREN encoding error require larger datasets for training and evaluation. Therefore, we train 300,000 SIRENs on a set of 100,000 images from the the MSCOCO dataset, all downsampled and center-cropped to 512x512 pixels.

For each of these 300,000 SIRENs, we record their hyperparameters, training loss curves, and highest-PSNR weights after 20,000 steps of training. The primary dependent variable of interest throughout this paper is the maximum PSNR value achieved during training. These 300,000 SIRENs are divided into two datasets. 100,000 of these SIRENs all share the same architecture, and only vary by the image that they encode. We call this collection of SIRENs the \textit{single-architecture dataset}. The remaining 200,000 comprise the \textit{many-architecture dataset}, and they are trained across a range of architectural hyperparameters. In total, these 300,000 SIRENs took roughly 12,000 GPU hours to train. Sections \ref{sec:single-arch-dataset} and \ref{sec:many-arch-dataset} describe these two sets of SIRENs in more detail.

\subsection{The Single-Architecture Dataset}
\label{sec:single-arch-dataset}

We train 100,000 SIRENs on the 100,000 images in our random sampling of the MSCOCO dataset. Each image is downsampled to 224x224 pixels and then used to train a SIREN with 8 layers, 32 hidden units per layer, and a $\omega_{0}$ of 15. If we store the SIREN parameters at 16-bit precision, this corresponds to a 0.9 bpp representation of each image. The PSNRs of the resulting images have a mean of \valSingleArchMean{} dB, a standard deviation of \valSingleArchStd{} dB, and are Gamma-distributed.

\subsection{The Many-Architecture Dataset}
\label{sec:many-arch-dataset}

We also train a collection of SIRENs over a range of hyperparameter values. For this group of SIRENs, we select 200,000 random samples of training parameters from the following range of options:

\begin{itemize}
    \item \textbf{Target image} - All images were chosen from our subset of \textbf{100,000 photographs} randomly sampled from the MSCOCO dataset. The training/validation/tests splits are done such that no image appears in two splits.
    \item \textbf{Image size} - We train our SIRENs on square images \textbf{between 112 and 512 pixels} in width.
    \item \textbf{Network Depth} - Following \cite{dupont2021coin}, we sweep through networks \textbf{between 2 and 12 layers deep}. We notice a dropoff in performance at both ends of this distribution, suggesting we have swept the useful range.
    \item \textbf{Network width} - Each SIREN we train has a uniform ``width,'' i.e. a uniform number of hidden units per layer. We sample the number of hidden units such that the SIRENs' compression ratios follow a log-uniform distribution \textbf{between 0.5 and 9 bits per pixel} (bpp).
    \item $\bm{\omega_{0}}$ - This is an important SIREN hyperparameter which determines the spatial frequency of the activations which come out of the first layer. \cite{sitzmann2020implicit} found that a $\omega_{0}$ of 30 worked well for their applications.  Consistent with findings from previous works \citep{Yuce_2022_CVPR, benbarka2022seeing, zhu2024disorder}, we find that  the optimal choice of $\omega_{0}$ varies with both the image size and image content. In general, larger images with more high frequency detail benefit from a higher $\omega_{0}$. Therefore, we randomly sample a value $\gamma = \frac{\omega_0}{\textit{image size}}$ from a log-uniform distribution between 0.02 and 0.12, and set $\omega_{0} = \gamma \times \textit{image size}$. Like the number of layers, this distribution appears to span the optimal range of $\omega_0$ values.
\end{itemize}

Two important parameters which we do not vary in this paper are 1) number of training steps and 2) learning rate. We find that across all of our networks, a learning rate of 0.001 is near-optimal. We also find that our SIREN networks show diminishing returns in PSNR after around 20,000 training steps. Indeed, during the last 1,000 steps, the median SIREN in our dataset improved by only \valLastThousandStepsMedianImprovement{} dB PSNR, and the 99\% of networks improved by less than \valLastThousandStepsTailImprovement{} dB. Therefore, we fix the number of training steps to 20,000 for all our networks. 

Because this random sampling includes many suboptimal architecture choices, the PSNR of the many-architecture dataset trends lower than in the single-architecture dataset. The average PSNR of this group is \valMultiArchMean{} dB, the standard deviation is \valMultiArchStd{} dB. Figure \ref{fig:psnr_distributions} in the Appendix shows the gamma-distributed histograms of PSNR values for the single- and many-architecture datasets.

\subsection{Metrics}

The main objective of our paper is to predict the maximum PSNR that a SIREN will obtain on the image it encodes, given the initial training conditions. Our analysis will use a few basic metrics:

\begin{itemize}
    \item \textbf{Maximum PSNR}: When training SIRENs, it is common practice to use the version of the model that achieved the best training loss, instead of the final model. We follow this practice and focus on the maximum PSNR seen during training as the dependent variable of interest. Unless otherwise stated, ``PSNR'' or ``encoding error'' refers to this maximum PSNR value.
    
    \item \textbf{RMSE / $R^2$ Coefficient / Explained Variance}: We also report the accuracy of our predictive models in terms of Root Mean Squared error (RSME), and explained variance (EV). Explained variance is defined as:

    \begin{equation}
    EV = 1 - \frac{Var[Y - \hat{Y}]}{Var[Y]} 
    \end{equation}
    
    This metric has a maximum value of 1 and is often expressed as a percentage. In the case of linear regression, explained variance is equivalent to the coefficient of determination, $R^2$. In such cases we may refer to the two concepts interchangeably. For example, when we say that a model explains 95\% of the variance in encoding error, this means that the $R^2$ coefficient between the predicted and actual encoding error is 0.95.

    \item \textbf{Irreducible Error}: Our predictive models do not account for all factors which affect the final SIREN PSNR. For example, the final PSNR reached depends on the random seed used during training, which is not an input to our models. Therefore perfect prediction is impossible. When trying to predict a variable Y (here, PSNR) from a variable X, the irreducible error $\epsilon$, i.e. the minimum possible RMSE that any predictive model can reach is:

    \begin{equation}
    \epsilon = \sqrt{\underset{x \sim X}{\mathbb{E}} Var[Y | X=x]}
    \end{equation}

    In our setting, X represents the SIREN’s training inputs: width, depth, $\omega_0$ and the target image. 

    To estimate this irreducible error, we sample 10,000 training inputs $x_1, x_2, \ldots, x_N \sim X$ from each of our SIREN datasets. For each input $x_i$, we sample two PSNRs $y_{i,1},y_{i,2} \sim P(Y|X=x_i)$. We then estimate the irreducible error using the following formula (see Appendix \ref{sec:RMSE} for a derivation):

    \begin{equation}
    \epsilon = \sqrt{\frac{1}{2N}\sum_{i=1}^{N} (y_{i,1} - y_{i,2})^2}
    \label{eq:RMSE}
    \end{equation}
    
\end{itemize}

\section{Experiments}
\label{sec:experiments}

In this project, we seek to understand how SIREN encoding error is affected by a variety of experimental variables. In addition to training and evaluating our encoding error prediction models, we perform a variety of controlled experiments, meant to isolate those experimental variables. This section is organized as follows:

\begin{itemize}
    \item Sections \ref{sec:random_variation} through \ref{sec:jpeg} provide context for understanding the SIREN encoding error landscape.
      \begin{itemize}
          \item Section \ref{sec:random_variation}
explores the irreducible error of our prediction problem due to random variation in Siren PSNR. 

        \item Section \ref{sec:power-laws} offers a useful rule of thumb for how encoding error varies with model size. 
        
        \item Sections \ref{sec:extrapolating-psnr}
 and \ref{sec:jpeg} present simple 1-variable linear regression models for predicting encoding error. We will use these models as a baseline comparison to our more sophisticated encoding error prediction networks.
      \end{itemize}
    
    \item In Section \ref{sec:GPR} we use Gaussian process regression to predict encoding error from the SIREN hyperparameters and image features.
    
    \item Section \ref{sec:predictive_models} reports the accuracy of our predictive models.
    \item Finally, Section \ref{sec:hyperparam-search} demonstrates a practical application of these predictive models. 
\end{itemize}
 
 \subsection{Random Variation in SIREN Encoding Error}
\label{sec:random_variation}

We find that there is significant random variation in what encoding error a given SIREN will reach, based solely on its random initialization. We quantify this random variation with an estimate of the standard deviation in PSNR among SIRENs which have been trained identically except for their random initialization. For the 10-layer-deep, 28-neuron-wide SIRENs trained on the Kodak dataset from COIN \citep{dupont2021coin}, the random variation is approximately $\pm$  \valKodakCoinIrrErr{} dB.

The magnitude of this random variation decreases with network width, and can be largely attributed to the random initialization of the first layer. The consequence is that very narrow, deep SIRENs like the ones used by COIN, are the most affected, but this effect can be ameliorated by holding the first layer fixed, or by using a meta-learned initialization. (See Appendix \ref{sec:random_variation_appendix} for further explanation.)

This random variation is important to keep in mind when comparing the performance of different SIREN architectures and training schemes. For example, \cite{dupont2021coin} conducted a small-scale hyperparameter sweep suggesting that 10 layers was an optimal depth for their networks . This result was obtained by training one network at each depth, and selecting the network with the best PSNR. We retrain these networks 10 times each with different random seeds, and then bootstrap a random sample of possible experimental outcomes which could have occurred. We found that 95\% of the time, their procedure selects optimal depths between 7 and 13 layers deep. This demonstrates that random variation can play an important role in the outcome of such small-scale experiments.

\subsection{SIREN Scaling Follows Power Laws}
\label{sec:power-laws}

\begin{wrapfigure}[14]{r}{0.5\textwidth}
  \centering
  \vspace{-5em}
  \includegraphics[width=\linewidth]{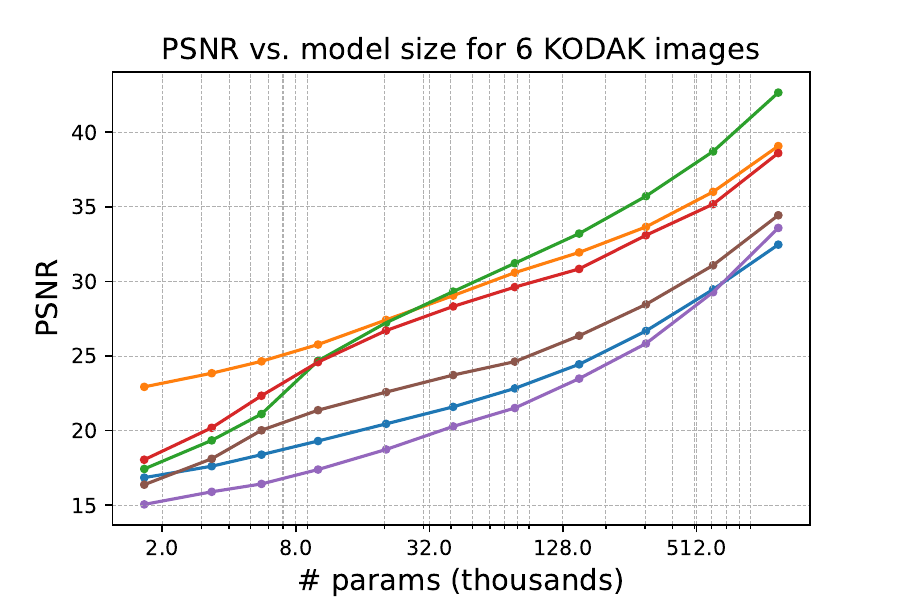}
  \vspace{-1.5em}
  \caption{Scaling curves for SIRENs trained on the first six images in the Kodak dataset. Each line represents the scaling curve for a different image.}
  \label{fig:power-laws}
\end{wrapfigure}

As a rule of thumb, we find that each doubling of the network size improves PSNR by about 2 dB. The literature on scaling laws posits a theoretical basis for this: 
\cite{bahri2021explaining} present a simple model of neural networks trained with large amounts of data and limited parameters, which they refer to as the \textit{resolution-limited} regime. According to their model, MSE loss $L$ should decrease with the number of parameters $P$ according to a power law $L(P) = \Omega(P^{-2/d})$, where $\Omega$ represents a lower bound and $d$ is the implicit dimension of the data manifold.

Since PSNR is proportional to the log of MSE, this law implies that the PSNR should grow linearly with $\log(P)$. Figure \ref{fig:power-laws} shows PSNR vs. $\log(P)$ for SIRENs trained on 6 of the 24 Kodak images, for networks with ten layers, and widths from 4 to 128 neurons per layer. For a limited range of network widths, this linear relationship holds up quite well.

However, the exponent of this power law appears to be contingent on our experimental setup. Much smaller or larger SIRENs, or SIRENs trained on very different data, will have different scaling curves. For example, notice that the image represented by the green line follows a steeper power law exponent than the other five images. And as we zoom out to consider networks which are orders of magnitude larger or smaller than the original COIN network, the relationship between PSNR and $\log(P)$ becomes nonlinear.

These power laws allow us to make local predictions about how the encoding error will change with the SIREN width. But to predict encoding error across other variables, such as the target image content, we will need a model which depends on those inputs as well.

\subsection{Extrapolating from Past PSNR to Future PSNR}
\label{sec:extrapolating-psnr}
Perhaps one of the simplest methods to predict the encoding error a SIREN will reach after $n$ training steps is to train the SIREN for $m < n$ steps, observe its encoding error, and extrapolate what the error will be at training step $n$, using linear regression. For example, the COIN networks are trained on the Kodak dataset for 50,000 steps each. The PSNR of each SIREN after 10,000 training steps explains 99.7\% of the variance in PSNR after 50,000 steps. Table \ref{tab:extrapolating-psnr} shows the explained variance (i.e. $R^2$ coefficient) you can reach with this prediction method for each of our datasets.

\begin{table}[]
\centering
\caption{Explained variance (EV) in PSNR after $n$ training steps from the PSNR after $m$ training steps, as calculated by $R^2$ score. For example, the PSNR after 1,000 steps explains 95.8\% of the variance in PSNR after 50,000 steps for the SIRENs trained with the images and architecture from COIN \citep{dupont2021coin}.}
\begin{tabular}{lcccc}
\toprule
 & & \multicolumn{3}{c}{EV (\%) after $m$ steps $\uparrow$} \\
\cmidrule(r){3-5}
% TODO: add a column for the difference in average PSNR between m steps and n steps.
Dataset & $n$ steps & 100 & 1,000 & 10,000 \\
\midrule
COIN & 50,000 & 2.5 & 95.8 & 99.7 \\
single-arch. & 20,000 & 25.8 & 93.0 & 99.8 \\
multi-arch. & 20,000 & 20.7 & 90.9 & 99.9 \\
\bottomrule
\end{tabular}
\label{tab:extrapolating-psnr}
\end{table}

We would like to predict the encoding error to a high accuracy in orders of magnitude less time than it takes to train the network. Table \ref{tab:extrapolating-psnr} shows that this simple extrapolation method is inadequate to achieve this goal. By step 10,000, these SIRENs are approaching their asymptotic behavior of sharply diminishing returns in PSNR per training iteration. In this regime, there is a strong correlation with the final PSNR, but only because the PSNR doesn't change very much in the remaining training steps: From step 10,000 to step 50,000, the COIN networks only improve by an average of 0.4 PSNR. We would be better off predicting the final PSNR using an equivalent JPEG compression, as explained in Section \ref{sec:jpeg}. In the next section, we will see that we can do better by solving this regression problem using neural nets.

\subsection{Comparing SIREN and JPEG Compression}
\label{sec:jpeg}

By storing trainable weights at 16-bit precision, \cite{dupont2021coin} use SIRENs as a compressed image representation competitive with JPEG2000 at low bitrates. We find that there is a strong correlation between how well JPEG and SIREN compression can fit a given image. This suggests a simple technique for predicting SIREN encoding error: find the encoding error of a SIREN architecture on a few images, and then use a standard zeroth-order numerical optimizer (e.g. scipy.optimize) to find the corresponding JPEG compression ratio which maximises the correlation between JPEG and SIREN encoding error.

\begin{figure}[t]
    \centering
    \subfloat[\label{fig:coin-vs-jpeg-kodak}]{%
        \includegraphics[width=0.25\textwidth]{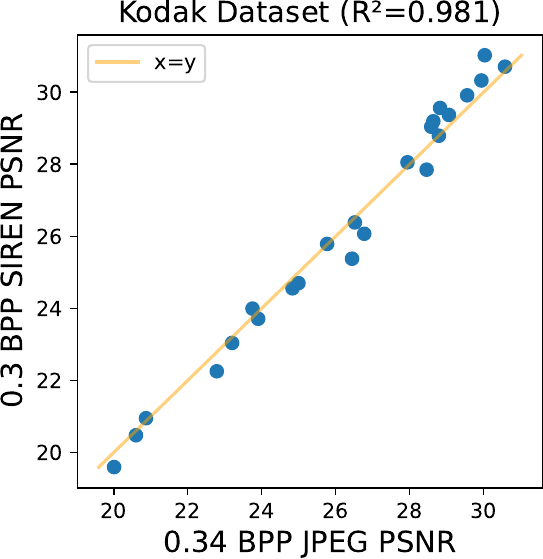}}
    \hfill
    \subfloat[\label{fig:coin-vs-jpeg-mscoco}]{%
        \includegraphics[width=0.26\textwidth]{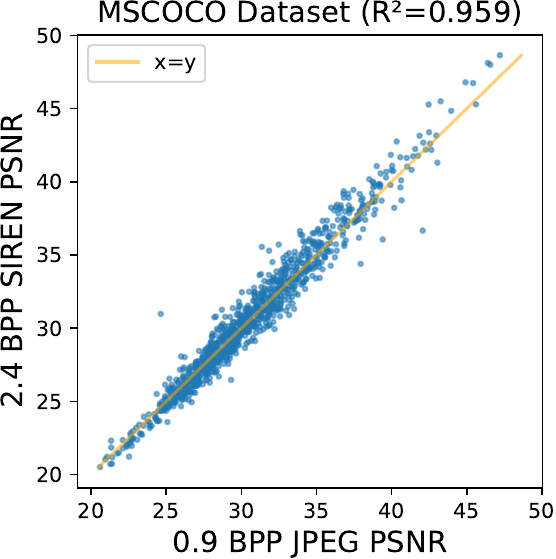}}
    \hfill
    \subfloat[\label{fig:coin-vs-jpeg-mscoco-multi}]{%
        \includegraphics[width=0.44\textwidth]{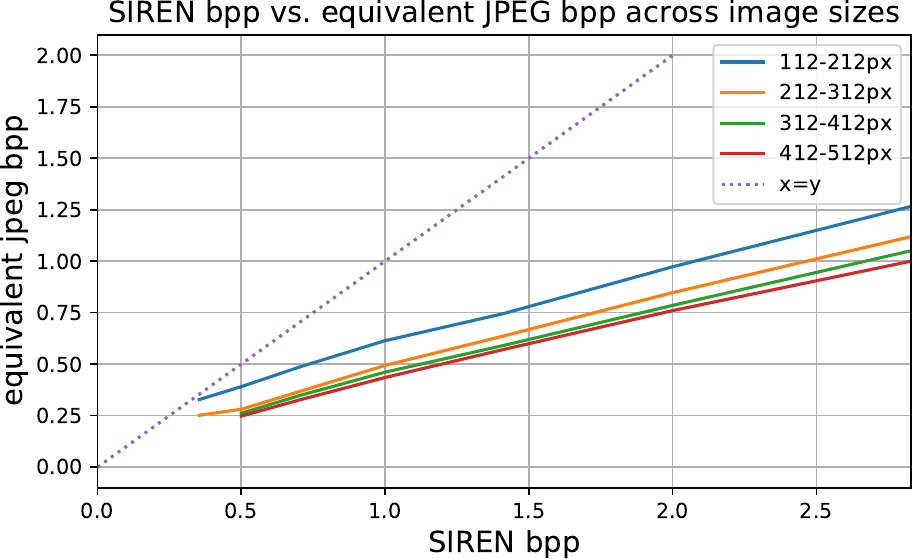}}

    \caption{SIREN vs. JPEG representations. In (a) and (b), each data point represents an image, and the PSNR to which it can be compressed by JPEG and COIN respectively. In (c) we show which JPEG compression ratio reaches the same PSNR as COIN for a given image size and COIN compression ratio.}
    \label{fig:coin-vs-jpeg}
\end{figure}

Using the Kodak dataset and SIREN hyperparameters from \cite{dupont2021coin}, we train a SIREN on each image to obtain a 0.3 bits-per-pixel (bpp) compressed representation of each image. Using JPEG2000 and compressing to 0.34 bpp, we find that the JPEG's PSNR explains 98\% of the variance in SIREN's PSNR (See Figure \ref{fig:coin-vs-jpeg-kodak}).

This correlation is not quite as strong for our single-architecture SIREN dataset. Compressing these images to 0.9 bpp using JPEG2000 explains 96\% of the variance in SIREN PSNR. JPEG PSNR predicts the SIRENs' PSNR to within 1.0 dB RMSE. See Figure \ref{fig:coin-vs-jpeg-mscoco}.

    Comparing JPEG and SIREN compression for our many-architecture dataset is more complex. For the Kodak and single-architecture sets of SIRENs, we treated JPEG compression as a one-parameter predictive model of SIREN PSNR. But with the multiple-architecture dataset, each SIREN architecture is best predicted by a different JPEG compression ratio. Therefore, we predict the encoding errors using a Gaussian Process (GP) regression model using the following features: JPEG PSNR on the target image at compression ratios of 7, 25, and 100, and the three SIREN hyperparameters: width, depth, and w0i. Coincidentally, this regression model predicts the SIREN's PSNR to within 1.1 dB RMSE. For more details, see Section \ref{sec:GPR}.

Additionally, for each trained SIREN network in our dataset, we find the JPEG compression ratio that reaches the same PSNR, and then analyze the relationship between SIREN bpp and equivalent JPEG bpp. For each image size, we observe a linear relationship between the compression ratio of our optimal-hyperparameter SIREN networks (as measured in bits per pixel) and the equivalent JPEG compression ratio, as can be seen in Figure \ref{fig:coin-vs-jpeg-mscoco-multi}. Our single-architecture dataset mostly contains SIRENs at bpps which are not competitive with JPEG, but notice that extrapolating the lines to the left indicates that our SIRENs should outperform JPEG at between 0.1 and 0.3 bpp. This is consistent with Figure 2 of \citep{dupont2021coin}, which shows SIRENs outperforming JPEG compression below 0.3 bpp.

\subsection{Extrapolating training curves using the empirical NTK}
\label{sec:extrapolating-ntk}

In some settings, gradient descent of the empirical, finite-width NTK is known to be a good and efficient approximation for optimizing the neural network itself \citep{mohamadi2023fast}. \cite{zancato2020predicting} used this approximation to successfully predict the training times on fine-tuning tasks to within 20\% accuracy. With full-batch training, the MSE loss $\mathcal{L}$ of the NTK approximation after $t$ steps of training with a learning rate $\eta$ develops according to the following equation \citep{arora2019fine}:

\begin{equation}
    \mathcal{L}(y_t) = ||(I - \eta \Theta)^t (y - y_0)||_2^2
    \label{eq:ntk-sgd}
\end{equation}

where $\Theta$ is the NTK matrix, $y_0$ is the network's output at initialization, and $y$ is the target output.

Unfortunately, we did not find success in applying this approximation to our networks. Following the method from \cite{zancato2020predicting}, we approximate the empirical NTK of the network at initialization and used this to extrapolate the network's learning curve. For our problem, these NTK-based extrapolations saturate very early, predicting that the network will essentially learn a flat-colored image where each pixel just takes on the average color of the target image.

\begin{wrapfigure}[17]{r}{0.45\textwidth}
    \centering
    \vspace{-1em}
    \includegraphics[width=0.5\textwidth]{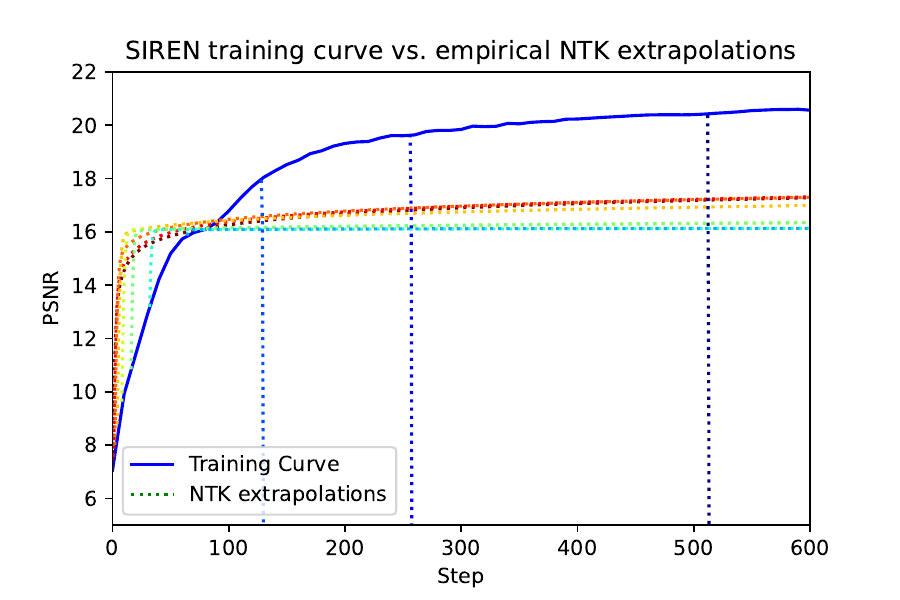}
    \caption{A typical example of a SIREN learning curve (thick blue line) vs. NTK-based extrapolations of that learning curve.}
    \label{fig:empirical-ntk-extrapolation}
\end{wrapfigure}

Others have found that even when the NTK at initialization is a poor proxy for network convergence, the empirical NTK can become a significantly better approximation after just a few steps of training \citep{kopitkov2020neural}. To test if this held for our problem, we took snapshots of our SIRENs' weights after $n=1,2,4,8...,2^{15}$ training steps, and used the NTK approximation from \cite{zancato2020predicting} to extrapolate the learning curves forward from that point onward.

Figure \ref{fig:empirical-ntk-extrapolation} shows the true training curve vs. empirical-NTK-extrapolated training curves starting from $n=1,2,4,8...,2^{15}$ training steps. It is clear that the SIRENs and their NTKs behave very differently. At first, the NTKs produced from snapshots of the network early in training follow the exact same dynamic described above: they learn only the DC component of the target function, and then saturate. Around 128 training steps, the learning rate of $\eta = 0.002$ leads to divergent behavior on the NTK model. This is due to the increasing eigenvalues of the kernel matrix $\Theta$; according to Equation \ref{eq:ntk-sgd}, SGD will begin to diverge when the largest eigenvalue of $\Theta$ exceeds $\frac{1}{\eta}$. Instead of decreasing, the loss increases exponentially with each training step.

As mentioned in Section \ref{sec:theoretical-models}, NTK theory is a good approximation for the behavior of \textit{overparameterized} networks. Our experiments here simply confirm that NTK-based predictions break down in the \textit{underparameterized} regime.

\subsection{Predicting Encoding Error using Gaussian Process Regression}
\label{sec:GPR}

In Neural Architecture Search and hyperparameter optimization problems, it is common to model the network's performance across different hyperparameters using Gaussian Process (GP) Regression \citep{snoek2012practical}. Our problem is very similar to a traditional neural architecture search problem, except that ours is also highly dependent on the image to encode. To apply GP regression successfully, we must extract appropriate image features.

Building on our findings from Section \ref{sec:jpeg}, we calculate the PSNR of JPEG2000 compression of our images at three different compression ratios: 7x, 25x, and 100x. We then fit a GP regression model on these PSNRs, plus our SIREN hyperparameters, to predict final encoding error. Due to the higher time complexity of GP regression compared to gradient descent, we cannot feasibly fit this regression model on the hundreds of thousands of samples in our training set. Therefore, our GP regression models are fit on just 1,000 random samples from the training set. This GP regression model obtains an RMSE of about 1 dB PSNR, as compared to the ~0.55 dB PSNR of our best prediction network. However, the GP model is more sample efficient, and the neural-network-based approach only outperforms it when given at least 5,000 training samples.

Additionally, we try fitting a GP model using features extracted using our pre-trained PSNR prediction network. We try extracting features from 1) the convolutional network which outputs a 512-dimensional image feature, and 2) layers 1, 2, and 3 from the multilayer perceptron (MLP) regression head, each of which is a 256-dimensional feature. When using the features from layer 3, we are essentially replacing the linear regression head in the last layer of our neural network with GP regression. In this case, average RMSE remains about the same, whether using linear regression or GP regression.

For further implementation details, see Section \ref{sec:GPR-details}.

\begin{figure}
    \centering
    \includegraphics[width=\textwidth]{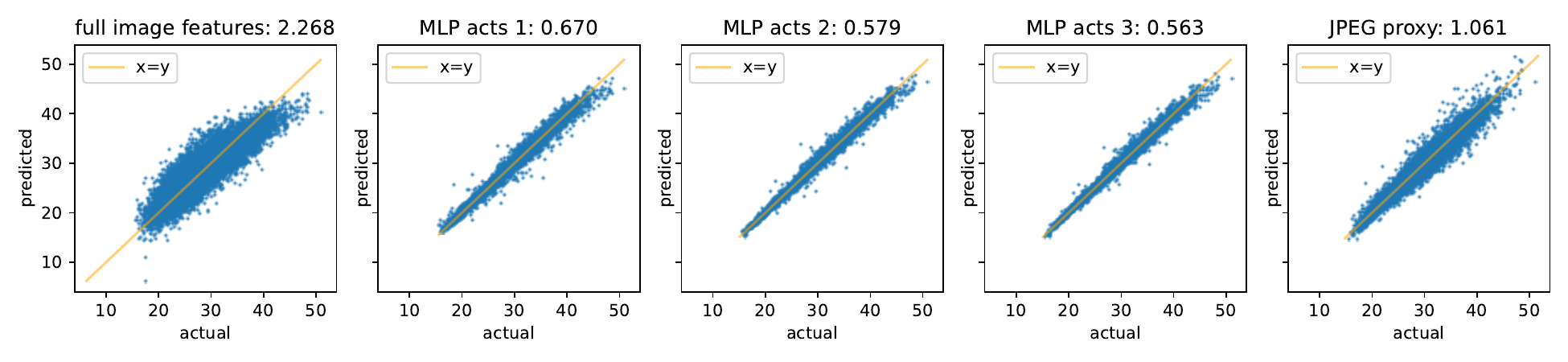}
    \caption{
        Actual vs. predicted PSNR for GP regression models trained on several different input features. From left to right- "full image features": 512-dimensional features extracted from each image using the CNN-component of our PSNR prediction network, concatenated with the SIREN hyperparameters, then "MLP acts": three different 256-dimensional features from the activations after each layer of the 3-layer MLP regression head, then "JPEG proxy": JPEG2000 proxy PSNRs derived by compressing the image to encode using compression ratios of 1:7, 1:25, and 1:100, concatenated with the SIREN hyperparameters. }
    \label{fig:enter-label}
\end{figure}

\subsection{Encoding Error Prediction Networks}
\label{sec:predictive_models}

Given an image to encode and SIREN hyperparameters, can we predict the encoding error the SIREN will attain? To answer this question, we trained neural networks on this regression problem for both our single- and many-architecture datasets. Our models are optimized to minimize the MSE of their PSNR predictions.

For the single-architecture dataset, all aspects of encoding besides the target image are held constant. We train a convolutional network to take in a target image, and predict what PSNR our SIREN will reach on that image (see Appendix \ref{sec:training-details} for details). We chose a variant of Normalization-Free Network from \cite{brock2021high} as our backbone CNN architecture, due to its high accuracy on the validation set. Appendix \ref{sec:arch-sweep} describes our backbone architecture selection procedure in more detail.

For the multiple-architecture dataset, we have several independent variables from which to predict PSNR: the training image, $\omega_{0}$, network width, and network depth. We feed all of these variables into a neural network whose architecture is depicted in Figure \ref{fig:overview}.

To train these models, we freeze the pretrained weights of the backbone image classifier and optimize the randomly-initialized ``regression head'' on the network for 10 epochs. Then we unfreeze the entire network for an additional 10 epochs. We use an 80/10/10 train/validation/test split, and select the version of the network which obtained the best validation set accuracy during training.

As a performance baseline, Section \ref{sec:jpeg} showed that we can predict SIREN PSNR from JPEG compression PSNR with over $94\%$ explained variance. Section \ref{sec:extrapolating-psnr} explores a simple method for predicting SIREN encoding error by only training the SIREN for a small number of iterations, and then extrapolating.

Our encoding error prediction networks outperform these simple baselines. On the single-architecture dataset, our model predicts the SIRENs' PSNR to within 0.30 dB RMSE, with an $R^2$ score of 0.996. The irreducible error of this prediction problem is approximately 0.21 dB PSNR. The many-architecture predictor is accurate to within 0.55 dB RMSE, to an $R^2$ score of 0.987. A portion of the higher error comes from a higher irreducible error of 0.27 dB RMSE, due to the inclusion of some suboptimal SIREN architectures which have a very high random variation in PSNR.

For the single-architecture dataset, the prediction error residuals are very nearly normally distributed. Interestingly, this is not the case for the many-architecture dataset, where the prediction error magnitudes are roughly exponentially distributed, with a median (i.e. ``half-life'') of 0.185 dB PSNR.

\begin{figure}
    \centering
    \subfloat[\label{fig:train_samples_single}]{%
        \includegraphics[width=0.5\textwidth]{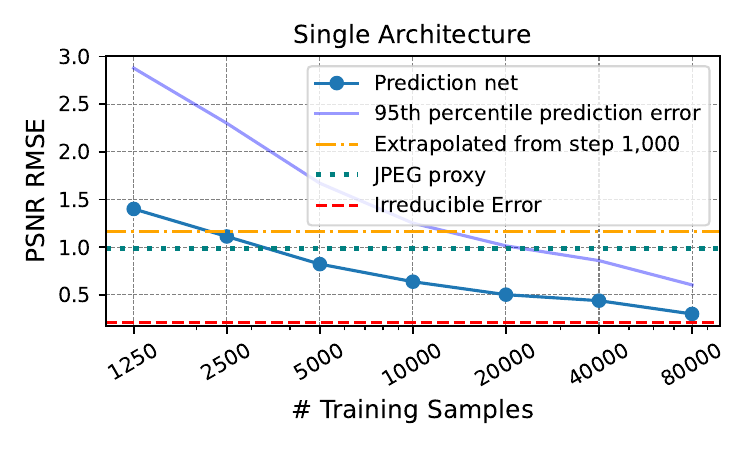}}
    \subfloat[\label{fig:single_scatter}]{%
    \raisebox{1em}{
        \includegraphics[width=0.275\textwidth]{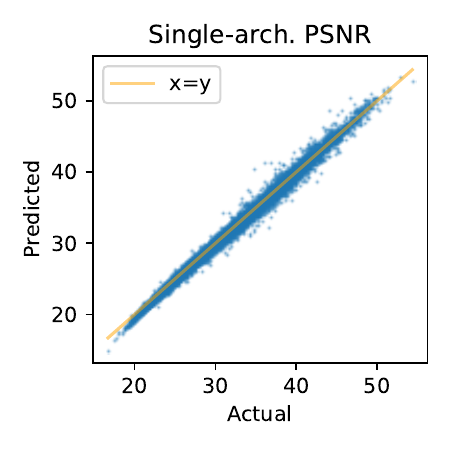}}}

    \vspace{-1.5em}
    \subfloat[\label{fig:train_samples_multiple}]{%
        \includegraphics[width=0.5\textwidth]{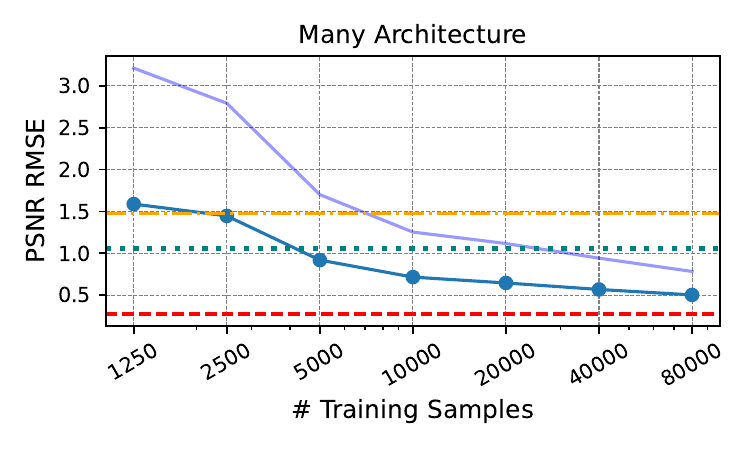}}
    \subfloat[\label{fig:many_scatter}]{%
        \raisebox{1em}{
        \includegraphics[width=0.275\textwidth]{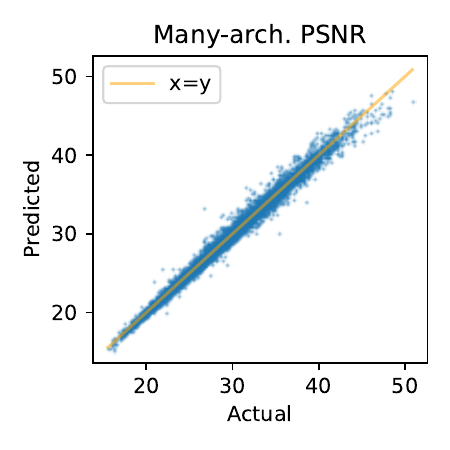}}}

    \caption{Number of training samples vs. test set error for the (a) single and (c) multiple architecture datasets. "95th percentile prediction error" is for the distribution of prediction net errors. Horizontal lines represent 1) the RMSE we get by linearly extrapolating the PSNR from training step 1,000, as described in Section \ref{sec:extrapolating-psnr}, 2) RMSE from predicting PSNR with a ``JPEG proxy model'' as depicted in Figure \ref{fig:coin-vs-jpeg}, and 3) the irreducible error of the regression problem, as described in Section \ref{sec:random_variation}. (b) and (d) are scatterplots of predicted vs. actual PSNR on the test sets, using the full training sets.}
    \label{fig:training-samples-vs-error}
\end{figure}

A key limitation our models is how much training data they require. The costliest step of producing these models is generating their training data, which requires training tens of thousands of SIRENs. Figure \ref{fig:training-samples-vs-error} reports the model's test accuracy for successive halvings of the training set size, compared to two far more sample-efficient baseline predictive models discussed above. The scaling curve is remarkably uniform, and suggests that even larger training sets would continue to improve test accuracy.

With a forward-pass time of 70 ms on a Titan X GPU, as opposed to an original encoding time of 2.5 minutes per network, this allows us to predict PSNR ~2,000$\times$ faster than we could by training the SIREN. For the many-architecture predictor, the speedup is even more significant: If we wish to evaluate multiple SIREN architectures against a single image, we can feed that image through the CNN-based feature extractor once, and then run only the small MLP regression head for each new model architecture. On our system, this regression head can process tens of thousands of inputs per second. In the next section, we will demonstrate how to take advantage of this speedup to perform an accelerated hyperparameter search.

\subsection{Accelerated Hyperparameter Search}
\label{sec:hyperparam-search}

By using our encoding error predictor as a fast proxy for SIREN training, we can accelerate any hyperparameter search method by more than 10,000x, as long as 1) the hyperparameters stay within the domain of our training data and 2) the objective function we are trying to optimize is a function of the variables of our predictive model.  In this section we explore two toy applications of our PSNR prediction network to hyperparameter search. 

First, given a SIREN size limit and an image to encode, we optimized the SIREN's width, depth, and $\omega_0$ to maximise the expected PSNR. However, we found that this approach did not confer a significant PSNR improvement over using a single SIREN architecture, selected for having the highest average PSNR across a test set of 100 images. This result held across many SIREN sizes and image sizes.

\begin{wrapfigure}[17]{r}{0.5\textwidth}
  \centering
  \vspace{-2em}
  \includegraphics[width=\linewidth]{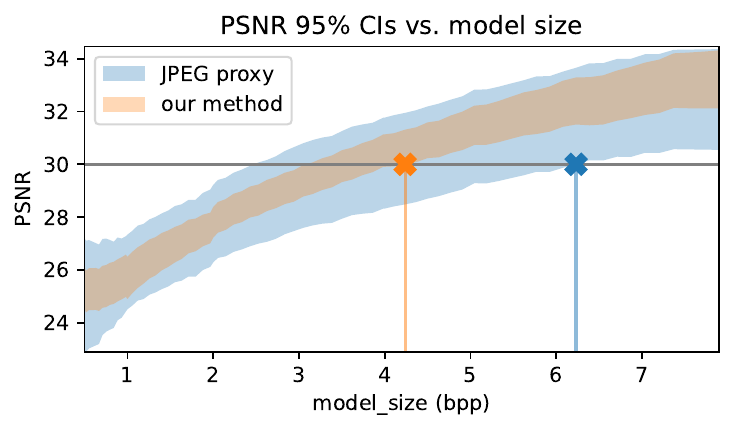}
  \caption{PSNR 95\% confidence intervals, as predicted by our predictive network (our method) and the JPEG proxy. The two vertical lines indicate the model sizes prescribed by these two predictive models to be 95\% sure the PSNR exceeds 30.}
  \label{fig:optimal_size_siren}
\end{wrapfigure}

Second, we chose a different optimization problem: given a target image, find the smallest SIREN network which will achieve at least 30 dB PSNR. However, since our PSNR predictions are approximate; we cannot \textit{guarantee} that the resulting PSNR will exceed 30. Assuming that the predictive model's error is normally distributed, we should aim for a predicted PSNR 2 standard deviations in error above 30 dB to be 95\% confident we will meet the requirement. In this context, the standard deviation of the error, or RMSE of our predictive model, plays a decisive role in determining how small of a SIREN we can use.

As a comparison to this hyperparameter search, we build a baseline using the JPEG-based PSNR predictor from Section \ref{sec:jpeg}. We find a sequence of 30 SIREN architectures of increasing size on the Pareto frontier of the network size / average PSNR tradeoff. Then we estimate the "effective JPEG compression ratio" for each of these SIREN architectures, as described in Section \ref{sec:jpeg}. Then, we use JPEG encoding error as a fast prediction of SIREN encoding error, and select the first model which we are 95\% confident will exceed 30 PSNR. For more details, see Appendix \ref{sec:hyperparam-search-details}.

As noted in section \ref{sec:predictive_models}, the JPEG proxy method is less accurate than our prediction network. In the context of this problem, the superior accuracy of our network-based predictive model translates directly into more efficient SIREN selection. Figure \ref{fig:optimal_size_siren} illustrates this process. The two predictive models, JPEG compression and the PSNR predictor, provide different 95\% confidence intervals in the PSNR that the optimal model of a given size will reach on a given image. A tighter confidence interval allows us to select a smaller model closer to that threshold. In our case, using the PSNR predictor allows us to select models which are on average 12\% smaller than those selected using the JPEG proxy method.

\section{Related Work}
\label{sec:related_work}

\subsection{Implicit Neural Representations}

The field of implicit neural representations, or INRs, is very broad, and in this section we will mention just a few of the many alternative coordinate network architectures. \cite{tancik2020fourier} showed that using random Fourier features as a positional encoding significantly improved the accuracy of implicit neural representations. SIRENs \citep{sitzmann2020implicit} extend this idea by using sine-wave activations, effectively turning the first layer of the network into a trainable positional encoding layer. Since then, many other positional encodings and activation functions for INRs have been proposed: \cite{ramasinghe2022beyond} present a unified framework for reasoning about the influence of different activation functions, and show that Gaussian activations are more stable to random initialization than SIRENs. \cite{Saragadam_2023_CVPR} propose a wavelet-based activation function which further improves the accuracy of implicit neural representations. FINER \citep{liu2023finer} uses a variable-frequency activation function to further increase the performance of INRs. There have been many more proposed schemes for INRs, such as hash-based methods \citep{muller2022instant, zhu2024disorder}, or multiplicative filter networks \citep{fathony2020multiplicative}, but SIRENs remain the most commonly used baseline, and this is why we choose them as our subject of study.

\subsection{Compression with Implicit Neural Representations}

INR-based data compression is the application most directly relevant to our research. Since the introduction of the SIREN network by \cite{sitzmann2020implicit}, researchers have sought to use SIRENs as compressed representations for images. In 2021 \cite{dupont2021coin} showed that SIRENs can outperform JPEG compression at low bitrates. We use their COIN (COmpression with Implicit Neural representations) architecture and training setup as a baseline for our experiments. 

Since then, several schemes have been devised to improve a SIREN's compression capabilities. These schemes employ a combination of meta-learning \citep{dupont2022coin++, schwarz2022meta, lee2021meta,strumpler2022implicit}, weight pruning \citep{lee2021meta, ramirez2022_0}, and quantization \citep{gordon2023quantizing} to improve the rate-distortion curves of SIREN-based image compression. \cite{guo2023compression} extend this compression paradigm to Bayesian neural networks. All of these projects attempt to obtain the best-possible rate-distortion curves for some set of images, although the size and content of the images studied varies across papers. \cite{gao2023sinco} propose a new loss term to enforce structural consistency between the reconstructed and ground truth images based on their segmentation maps. Almost all of these works build upon the SIREN architecture; therefore we believe that our observations about SIREN-based image compression are broadly relevant to this whole body of work.

\cite{dupont2022coin++} show that INRs can be used to compress many other data types, including sound, MRI, and weather data. Video is an especially common and data-intensive modality which has received special attention: while \cite{sitzmann2020implicit} and subsequent papers such as \cite{mehta2021modulated} applied SIRENs to relatively small (e.g. 7x224x224) videos, more specialized INR architectures such as NeRV \citep{chen2021nerv}, E-NERV \cite{li2022nerv}, NVP \cite{kim2022scalable}, and many others \citep{rho2022neural, zhang2021implicit, he2023towards, chen2022cnerv, Maiya_2023_CVPR}, have reached performance comparable with conventional video compression methods such as H.264 \citep{wiegand2003overview} and HEVC \citep{sullivan2012overview}. While outside the scope of the current work, our encoding error prediction method could be extended to these other modalities.

\subsection{Theoretical Models of Neural Networks}
\label{sec:theoretical-models}

Our primary goal is to develop a predictive model of how well a SIREN can encode a target image. While our methods are data-driven and empirical, others works have developed relevant theoretical models of network training dynamics, which may be predictive of SIREN performance. In particular, Neural Tangent Kernel (NTK) theory \citep{jacot2018neural} offers one possible tool for modelling the performance of SIRENs. As networks become sufficiently overparameterized, they enter a \textit{lazy training regime}, where their training is well-approximated by Taylor series expansion around their initialization \citep{atanasov2022onset}. Even outside of this lazy training regime, researchers have successfully applied NTK theory to explain certain properties of coordinate networks. For example, \cite{tancik2020fourier} use NTK theory to explain the advantage conferred by using a random Fourier feature (RFF) positional encoding for coordinate networks. As they show, applying NTK theory to neural networks with an RFF positional encoding naturally leads to a frequency-based perspective of the network's behavior.

There are many works based upon or inspired by a frequency-based perspective on implicit neural representations. \cite{ronen2019convergence} perform a more extensive analysis of the convergence rate of neural networks to functions of different frequencies, and confirm their theory matches observations for a few circumstances. \cite{Yuce_2022_CVPR} employ a different mathematical approach to reach similar conclusions about the spectral bias of the fourier-feature networks and SIRENs. \cite{ramasinghe2022beyond} analyze the choice of activation functions in implicit neural representations \cite{benbarka2022seeing} and \cite{zhu2024disorder} also introduce methods motivated by Fourier analysis of implicit neural representations.

However, none of these works precisely predict the loss curves of \textit{underparameterized} networks, which lack enough parameters to fully fit their training data. While some general arguments from NTK theory still apply, the approximation is too rough to make the kind of fine-grained PSNR predictions we seek.

For example, in \textit{Predicting Training Time Without Training}, \cite{zancato2020predicting} use neural tangent kernels to predict the fine-tuning loss curves of pretrained image classification networks. On the surface, their problem is extremely similar to ours. But their networks operate in the lazy training regime, while ours do not. Due to this difference, we found that their method does not transfer well to our setting. See Appendix \ref{sec:extrapolating-ntk} for more details.

\cite{bahri2021explaining} propose a theoretical model which is more applicable to our setting. They identify four regimes of neural network training dynamics, including a \textit{resolution-limited regime}, where the number of training samples far exceeds the number of parameters. They argue that the model's loss will decrease as a power law of the number of trainable parameters. In Section \ref{sec:power-laws}, we show that this model is in closer accordance with our observations. However, this power law only tells us how the encoding error will vary with model size, with everything else held constant. We are specifically interested in how PSNR varies with the \textit{training data}, and here their model offers no guidance.

\subsection{Meta-Learning Implicit Neural Representations}

Meta-learning, such as MAML \citep{finn2017model}, optimizes the initial weights of a neural network to converge quickly to any one of a set of target functions. Meta-learning can dramatically reduce the amount of time needed to fit a SIREN to a given signal \citep{Tancik_2021_CVPR, dupont2022data, lee2021meta}. Because meta-learning optimizes the initial weights of the network, it implicitly holds the network's hyperparameters, such as its width and depth, fixed. However, our encoding error prediction model works over a continuous range of possible hyperparameters.

\subsection{Neural Architecture Search}
\label{sec:NAS}

Our problem is very similar to the problem of neural architecture search (NAS). In both cases, the challenge is to predict the performance of a given neural network architecture before training it. NAS is a mature field with a great variety of established solutions. Here, we connect our work to just a few sample papers from this broad field.

We wish to predict how the network's performance varies with both architecture and \textit{training task}, i.e the image to encode. However, as the name implies, NAS focuses primarily on variations in architecture, while \textit{task-aware} NAS methods are rarer. SMAC \citep{hutter2011sequential} can predict model performance from a mixture of both hyperparameters and task meta-features. However, this method is based on a bayesian optimization algorithm which becomes slow for large sample sizes, and is not well-suited for highly complex, high-dimensional features such as images. \cite{istrate2019tapas} introduce TAPAS, a predictive model of neural network accuracy which also considers the difficulty of the training data using a method from \cite{scheidegger2021efficient}. \cite{kokiopoulou2020task} learn an embedding space for image classification datasets, which are then used to make task-aware predictions of neural architecture performance. At a high level, our problem and approach are very similar to theirs. The key difference lies in the choice of domain: we study SIRENs trained on single images, as opposed to classification networks trained on labeled image datasets.

\subsection{Predicting NeRF Parameters}

While we train a neural network to predict a coordinate network's encoding error on a target signal, other works go a step further, training ``hypernetworks'' which generate the coordinate network to model the signal. For example, \cite{lorraine2023att3d} and \cite{xie2024latte3d} produce NeRFs from textual descriptions. \cite{hong2023lrm} and \cite{pavllo2023shape} generate Signed Distance Functions (SDFs) and NeRFs from monocular views of an object. 

We place ``hypernetworks'' in quotes here because these networks do not directly output the NeRF's parameters, as is common for true hypernetworks \citep{ha2016hypernetworks}; instead, their output modulates the activity of a pretrained NeRF somehow, either by directly modulating the activations \citep{lorraine2023att3d}, or by generating triplanes which are used as inputs to the NeRF/SDF \citep{hong2023lrm, pavllo2023shape}. As for neural architecture search, there are many more works on this topic than we have cited; we are simply pointing out a few sample projects. Our dataset, which contains the trained weights of 300,000 SIRENs, may be of interest to those who study coordinate hypernetworks.

\section{Conclusions}

We have observed several important trends in SIREN encoding error. First, SIRENs have significant random variation in performance between random initializations, which is mostly explained by the first layer. This effect gets more pronounced as the networks become narrower. Pre-selecting a high-quality positional encoding layer instead of sampling one randomly appears to be an easy win for improving the encoding error of a narrow SIREN. We also find that SIREN PSNR correlates strongly with JPEG PSNR. SIRENs outperform JPEGs when the representation is very small: representing a small image or to a very low bpp.

We have shown that a relatively standard deep-learning architecture can solve a novel regression problem: predicting the encoding error a SIREN will reach on a given image. Our encoding error prediction networks compare favorably to three alternative approaches: 1) using JPEG compression loss as a proxy for SIREN loss, 2) linearly extrapolating future PSNR from a few steps of gradient descent and 3) NTK-based approximations of SIRENs in Section \ref{sec:extrapolating-ntk}. We show that this works for a single SIREN architecture, and also across a range of possible architectures. We can leverage such predictive models to simulate hyperparameter sweeps over the the SIREN architecture search space 10,000x faster than would otherwise be possible.

\section{Discussion}

The ``holy grail'' of this line of research is a theoretically well-founded model which can predict the training loss (here, encoding error) of underparameterized neural networks with high accuracy. Towards this aim, we suspect that our dataset of 300,000 SIRENs contains a lot of scientific value beyond the scope of this work. SIRENs trained to fit photographs using mean-squared-error loss are a promising ``model'' deep learning problem, in the same way that fruit flies are a common model organism. This model has several advantages:

\begin{itemize}
    \item \textbf{Architecturally Simple}: SIRENs are a simple variation on multi layer perceptrons, which are one of the most ubiquitous architectural building blocks in deep learning. 
    \item \textbf{Computationally Cheap}: In the context of our coordinate networks, a single photograph represents an entire training dataset. Both our SIRENs and their training data (photographs) have a low memory footprint, less than a megabyte each. Indeed it is this quality which has allowed us to publish a dataset of 300,000 of them.
    \item \textbf{Understandable}: Perhaps no naturally-occurring kind of dataset is more amenable to visualization than a photograph. The training samples, image pixels, are positioned on a regular grid, making them uniquely amenable to mathematical analyses such as the discrete Fourier transform. The loss function used to train our models, $L_2$ loss, is extremely well-studied and mathematically well-behaved.
\end{itemize}

Our 300,000 trained SIRENs provide a lot of empirical data about how simple neural networks learn to approximate functions. We hope that researchers who are interested in modeling the behavior of neural networks, especially networks in the \textit{resolution-limited regime} \citep{bahri2021explaining}, can evaluate their theories against our data. 

\bibliography{main}
\bibliographystyle{tmlr}
\appendix
 \section{Appendix}

\subsection{Gaussian Process Regression Implementation Details}
\label{sec:GPR-details}

All input features for our GP regression models---the neural activation vectors, JPEG2000 PSNR scores, and SIREN hyperparameters---were normalized to have a standard deviation of 1. Each GP Regression model was fitted on 1,000 images from the training set, and then evaluated using 10,000 samples from our test set.

All of our GP regression models use a composite kernel consisting of a constant kernel multiplied by a Radial Basis Function (RBF) kernel, plus a white kernel. The white kernel is intended to model the irreducible error in our data, see Section \ref{sec:random_variation}. We used standard GP Regression via Scikit-Learn's GaussianProcessRegressor class.

\subsection{Encoding Error Predictor Implementation Details}
\label{sec:training-details}

\subsubsection{Single Architecture Implementation Details}
To train our single-architecture encoding error predictor, we use the ADAM optimizer with a learning rate of 0.0001 and a batch size of 8. PSNR values are normalized to have a mean of 0 and a standard deviation of 1 for the training data. We start from a backbone pretrained classification network. (See Section \ref{sec:arch-sweep} for details.) We remove the classification head, and replace it with a ``regression head'', a fully-connected layer which outputs a single value: the PSNR prediction. We freeze all weights in the network except the regression head and perform 10 epochs of training, after which we unfreeze the entire network for an additional 10 epochs of training. Finally, we select network checkpoint from the epoch with the best validation set accuracy.

\subsubsection{Many Architecture Implementation Details}
The many-architecture encoding error predictor is slightly more complex, and has more implementation details to consider. 

First is the positional encoding, which encodes the SIREN hyperparameters (width, depth, $\omega_0$, and image size), into a vector in $\mathbb{R}^{80}$. We use the positional encoding scheme from \cite{mildenhall2021nerf}:

\begin{equation}
    \gamma(p) = (\sin(2^0 \pi p), \cos(2^0 \pi p),\dots,\sin(2^{10} \pi p), \cos(2^{10} \pi p))
\end{equation}

First, the hyperparameters are normalized to lie in the range (0,1). Log-uniformly distributed parameters $w$ and $\omega_i$ are re-scaled by taking their logarithm first, before normalization. For normalized width $w$, depth $d$, image size $s$, and $\omega_0$, the full positional encoding of the hyperparameters is:

\begin{equation}
    \Gamma(w, d, s, \omega_0) = (\gamma(w), \gamma(d), \gamma(s), \gamma(\omega_0))
\end{equation}

This positional encoding and the feature vector from the last layer of the classification network are concatenated and fed into the MLP regression head. In our experiments, we found that the performance of the multi-architecture classifier was not too sensitive to the width and depth of this MLP head, so we keep the network relatively small for performance reasons: 4 layers deep and 128 hidden units wide.

To train the multi-architecture encoding error predictor, we use the same optimizer, learning rate, and batch size as for the single-architecutre predictor. We start by freezing the weights of the pretrained classifier network, and training just the MLP regression head for 10 epochs. Then we unfreeze the classifier and train the entire network for 10 epochs, selecting the checkpoint with the epoch with the best validation accuracy.

\subsection{Choosing the PSNR Predictor Backbone Architecture}
\label{sec:arch-sweep}

Our single-architecture encoding error predictor is simply a fine-tuned image classification network. We used the PyTorch Image models library (TIMM) \citep{rw2019timm} to quickly sweep through hundreds of candidate classification networks to fine-tune on our task. Table \ref{tab:all-timm-models} shows the $R^2$ scores on the validation data from our initial sweep over many of the models in the TIMM library. Note that some details of training are not the same as in our final model, for example, these models were trained for fewer epochs. Table \ref{tab:best-timm-models} shows a more thorough evaluation on some of the best models identified in Table \ref{tab:all-timm-models}. Note that for this second sweep, we excluded some of the best-performing models from Table \ref{tab:all-timm-models} due to their slow training times. In the end, this sweep led us to choose ECA NFnet L0, a TIMM-specific variant of the NFNET architecture from \cite{brock2021high} as our final model architecture.

\begin{table}[ht]
\centering
\begin{tabular}{l r r}
     \toprule
     Model & Validation $R^2$ & Test $R^2$ \\
     \midrule
     eca nfnet L0 & 0.9956 & 0.9956\\
     nfnet L0 & 0.9954 & 0.9955 \\
     eca nfnet L1 & 0.9952 & 0.9953 \\
     coat lite mini &  0.9933 & 0.9936 \\
     nf regnet b1 & 0.9932 & 0.9933 \\
     convnextv2 large & 0.9923 & 0.9923 \\
     coat lite small & 0.9922 & 0.9924 \\
     \bottomrule
\end{tabular}
\caption{7 final architectures we tested before making a final decision to use nfnet for the rest of our experiments.}
\label{tab:best-timm-models}
\end{table}

\begin{longtable}{l r l r}
\caption{128 backbone classification architectures we tried from the Torch Image Models (TIMM) library \citep{rw2019timm}, on an early, smaller version of the single-architecture dataset. Ordered by the $R^2$ score they obtained on the validation set.}
\\
\textbf{model} & \textbf{$R^2$} & \textbf{model} & \textbf{$R^2$} \\ 
\hline
\endfirsthead
\textbf{model} & \textbf{$R^2$} & \textbf{model} & \textbf{$R^2$} \\ 
\hline
\endhead
eca\_nfnet\_l1 & 0.987 & eca\_nfnet\_l2 & 0.986 \\ 
convnext\_large & 0.984 & convnext\_base & 0.984 \\ 
nf\_regnet\_b1 & 0.983 & convnext\_base\_in22k & 0.981 \\ 
coat\_lite\_mini & 0.980 & convnext\_tiny & 0.980 \\ 
convnext\_xlarge\_in22ft1k & 0.979 & eca\_nfnet\_l0 & 0.978 \\ 
coat\_mini & 0.977 & convnext\_small & 0.976 \\ 
nfnet\_l0 & 0.975 & convnext\_base\_384\_in22ft1k & 0.974 \\ 
coat\_tiny & 0.973 & convnext\_base\_in22ft1k & 0.971 \\ 
deit\_tiny\_patch16\_224 & 0.967 & crossvit\_tiny\_240 & 0.965 \\ 
crossvit\_15\_dagger\_408 & 0.964 & coat\_lite\_small & 0.963 \\ 
crossvit\_small\_240 & 0.962 & crossvit\_15\_240 & 0.962 \\ 
crossvit\_9\_240 & 0.961 & deit\_base\_patch16\_384 & 0.958 \\ 
pit\_s\_224 & 0.956 & crossvit\_18\_dagger\_240 & 0.956 \\ 
crossvit\_9\_dagger\_240 & 0.956 & mixer\_b16\_224\_miil & 0.954 \\ 
pit\_b\_224 & 0.953 & crossvit\_base\_240 & 0.952 \\ 
jx\_nest\_base & 0.952 & jx\_nest\_small & 0.951 \\ 
cait\_xxs36\_224 & 0.950 & nf\_resnet50 & 0.950 \\ 
convnext\_xlarge\_in22k & 0.946 & crossvit\_15\_dagger\_240 & 0.946 \\ 
crossvit\_18\_240 & 0.945 & crossvit\_18\_dagger\_408 & 0.944 \\ 
coat\_lite\_tiny & 0.943 & resmlp\_12\_distilled\_224 & 0.939 \\ 
pit\_ti\_224 & 0.930 & deit\_base\_patch16\_224 & 0.929 \\ 
convnext\_large\_in22k & 0.928 & resnetv2\_50x1\_bit\_distilled & 0.927 \\ 
resnet50\_gn & 0.927 & gluon\_seresnext101\_32x4d & 0.926 \\ 
cait\_xxs24\_224 & 0.922 & resmlp\_36\_distilled\_224 & 0.921 \\ 
regnety\_016 & 0.921 & regnety\_006 & 0.918 \\ 
jx\_nest\_tiny & 0.916 & res2net50\_14w\_8s & 0.914 \\ 
convnext\_large\_in22ft1k & 0.913 & gluon\_resnext101\_32x4d & 0.912 \\ 
regnetz\_c16 & 0.909 & regnety\_002 & 0.908 \\ 
gluon\_seresnext50\_32x4d & 0.907 & gluon\_senet154 & 0.905 \\ 
eca\_botnext26ts\_256 & 0.904 & ese\_vovnet39b & 0.904 \\ 
resnest50d\_4s2x40d & 0.903 & regnety\_008 & 0.902 \\ 
regnety\_040 & 0.902 & resnetv2\_50x1\_bitm & 0.901 \\ 
ecaresnet26t & 0.899 & gernet\_m & 0.899 \\ 
resnetrs101 & 0.899 & halo2botnet50ts\_256 & 0.899 \\ 
regnetz\_d8 & 0.898 & regnetz\_e8 & 0.898 \\ 
gluon\_seresnext101\_64x4d & 0.898 & deit\_small\_patch16\_224 & 0.897 \\ 
regnetx\_004 & 0.892 & resnetrs200 & 0.892 \\ 
resnetrs50 & 0.890 & convit\_base & 0.889 \\ 
repvgg\_b0 & 0.889 & eca\_halonext26ts & 0.888 \\ 
regnetz\_d32 & 0.888 & eca\_resnet33ts & 0.888 \\ 
eca\_resnext26ts & 0.887 & regnety\_004 & 0.887 \\ 
seresnet152d & 0.885 & gcresnext26ts & 0.885 \\ 
resnet26 & 0.885 & regnetx\_032 & 0.884 \\ 
resnet152d & 0.883 & seresnext26d\_32x4d & 0.883 \\ 
pit\_xs\_224 & 0.883 & dpn68b & 0.881 \\ 
regnetx\_008 & 0.881 & regnetx\_016 & 0.879 \\ 
gluon\_resnet50\_v1c & 0.877 & resnet101d & 0.874 \\ 
resnetv2\_50x1\_bitm\_in21k & 0.874 & resnet50d & 0.873 \\ 
resnetblur50 & 0.872 & cait\_s24\_224 & 0.872 \\ 
resnet26t & 0.872 & regnetx\_120 & 0.871 \\ 
resnetv2\_101x1\_bitm & 0.871 & mixer\_b16\_224\_miil\_in21k & 0.870 \\ 
gcresnext50ts & 0.869 & lambda\_resnet26t & 0.868 \\ 
densenetblur121d & 0.867 & regnetx\_006 & 0.866 \\ 
gluon\_resnet152\_v1c & 0.866 & mixer\_l16\_224 & 0.865 \\ 
regnetx\_002 & 0.865 & resnest50d\_1s4x24d & 0.865 \\ 
gluon\_resnet101\_v1d & 0.865 & resnet26d & 0.865 \\ 
lambda\_resnet50ts & 0.865 & gmlp\_s16\_224 & 0.864 \\ 
gluon\_resnet50\_v1d & 0.864 & bat\_resnext26ts & 0.863 \\ 
gernet\_l & 0.861 & lamhalobotnet50ts\_256 & 0.861 \\ 
repvgg\_a2 & 0.860 & repvgg\_b1g4 & 0.860 \\ 
gluon\_resnext50\_32x4d & 0.858 & regnety\_032 & 0.857 \\ 
gcresnet50t & 0.857 & resnext26ts & 0.856 \\ 
regnetx\_080 & 0.855 & gluon\_resnet101\_v1c & 0.852 \\ 
regnety\_080 & 0.850 & resnet33ts & 0.848 \\ %\bottomrule
\hline
\label{tab:all-timm-models}

\end{longtable}

\newpage
\subsection{Input Ablation Study}
\label{sec:ablation-studies}

\begin{wraptable}[9]{r}{0.4\textwidth}
\centering
\vspace{-4em}
\begin{tabular}{@{}lrr@{}}
\toprule
\textbf{Removed Input} & \textbf{Error} $\downarrow$ & $\mathbf{R^2}$ $\uparrow$ \\ 
\midrule
\textit{No inputs removed} & 0.55 & 0.987 \\ 
Image Size & 0.71 & 0.980\\
$\omega_{0}$ & 1.25 & 0.939 \\
Number of Layers & 2.27 & 0.794\\ 
Layer Size & 2.52 & 0.746\\
All Hyperparameters & 2.90 & 0.668\\ 
Image & 4.11 & 0.317 \\
\bottomrule
\end{tabular}
\caption{Many-architecture prediction error with various removed inputs.}
\label{tab:input-ablations}
\end{wraptable}

Which inputs to our model are important for accurately predicting PSNR? Table \ref{tab:input-ablations} shows the test error of our multiple-architecture PSNR predictor with various combinations of inputs to the network removed. We see that removing any of the input features significantly degrades the accuracy of the prediction network, indicating that the final PSNR varies predictably with each of these inputs.

\newpage

\subsection{Accelerated Hyperparameter Search Implementation Details}
\label{sec:hyperparam-search-details}

In Section \ref{sec:hyperparam-search}, we look at the problem of selecting which of 30 SIREN architectures of progressively increasing size we should use to encode an image to at least 30 dB PSNR. We compared two methods, one based on the JPEG proxy model from section \ref{sec:jpeg}, and one based on our PSNR prediction network from Section \ref{sec:predictive_models}.

For the JPEG proxy model, we need an "effective JPEG compression ratio" for each of our 30 SIREN architectures, which are sized to compress the encoded image to between 0.5 and 8 BPP. We estimate these ratios by selecting 5 SIREN architectures that span the range of sizes (sizes which compress the image to 0.5, 1, 2, 4, and 8 bpp), and train these architectures on 30 images each. Then for each architecture, we use those 30 images to estimate an effective JPEG compression ratio, as described in Section \ref{sec:jpeg}. We also estimate the RMSE of our JPEG proxy models. The models' predicted PSNRs, and our estimated RMSEs allow us to estimate confidence intervals for the PSNR value, assuming normality. To get effective JPEG compression ratios and RMSEs for the remaining 25 models, we linearly interpolate those values from the 5 sample models.

We reuse the PSNRs of those same 30*5=150 SIRENs to estimate the RMSE of our PSNR prediction network for each SIREN architecture, again, using linear interpolation.

\newpage
\subsection{Random Variation in SIREN PSNR (contd.)}
\label{sec:random_variation_appendix}

\begin{wrapfigure}[13]{r}{0.4\textwidth}
  \centering
  \vspace{-5em}
  \includegraphics[width=\linewidth]{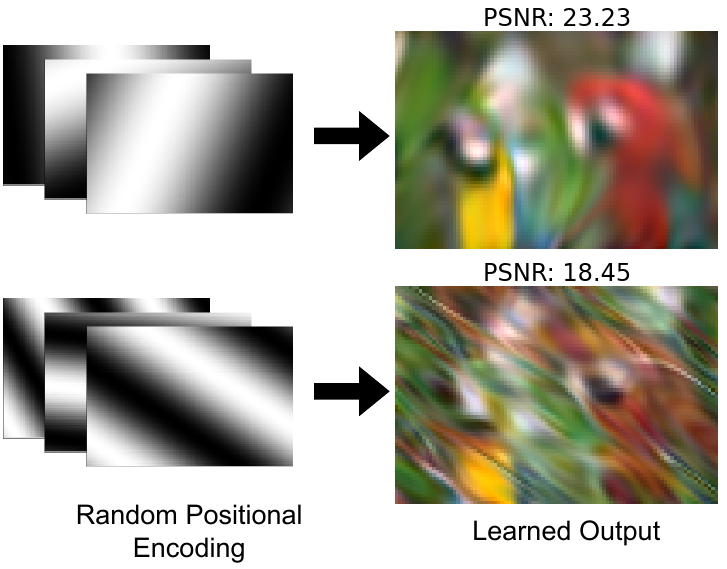}
  \vspace{-1.5em}
  \caption{Output of two small SIRENs with randomly-initialized 3-channel positional encodings.}
  \label{fig:pe-variation}
\end{wrapfigure}

Despite comprising less than 1\% of the network's learnable parameters, the first layer is responsible for roughly \valKodakCoinFirstLayerIrrErrEV{}\% of COIN's variance in final PSNR due to random initialization. By initializing the first layer in the same way each time instead of randomly, we reduce the standard deviation in PSNR among SIRENs using Dupont \etal's method from \valKodakCoinIrrErr{} to \valKodakCoinIrrErrWithFixedPE{} PSNR. 

\begin{figure}
    \centering
    \includegraphics[width=0.8\textwidth]{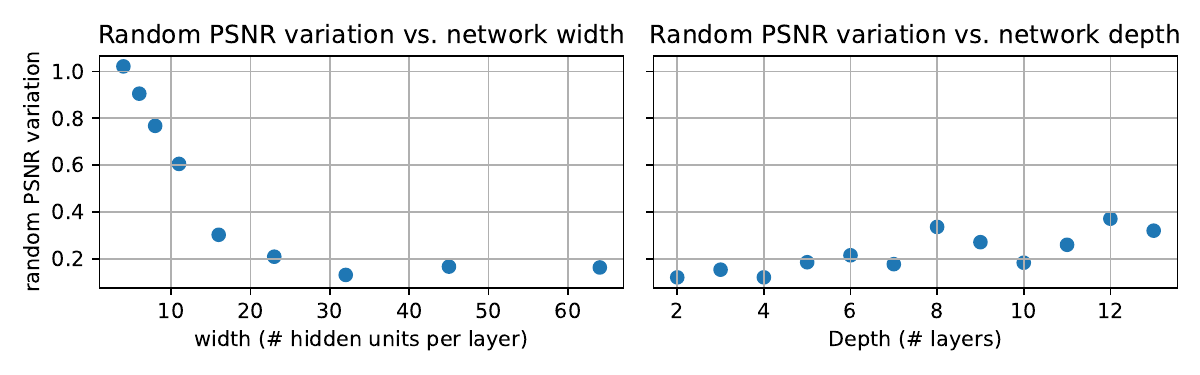}
    \caption{Effect of SIREN width and depth on the random variation in PSNR. Both sweeps begin from a baseline architecture of 10 layers, 32 hidden units per layer. In the width sweep, the depth is held constant, and vice-versa. For each choice of width/height, 10 SIRENs are trained on the same image, and the standard deviation in PSNR is plotted. Notice how the random variation in PSNR increases dramatically for narrow networks.}
    \label{fig:variation-vs-width-vs-depth}
\end{figure}

This random variation is inversely correlated with network width (See Figure \ref{fig:variation-vs-width-vs-depth}). When SIRENs are wide, they take larger random samples of the space of random Fourier features, and these large random samples converge to a similar distribution. But for very narrow networks, such as the ones used by COIN, the positional encoding consists of only a few randomly sampled sine waves, so there is significant variation in the quality of the positional encoding. Figure \ref{fig:pe-variation} illustrates this principle for very small SIRENs with a width of just 3 hidden units in the first layer. In this extreme case, the effect of a poorly initialized PE layer is especially clear.

We can improve SIREN performance by selecting a good first layer, instead of choosing one at random. We train 100 randomly initialized COIN networks on each of the 24 Kodak images, for a total of 2,400 trained networks. For each of the 24 images, we take the first layer from the SIREN with the best final PSNR, and train 24 new SIRENs, one for each image in the Kodak dataset, which are initialized with this layer. This gives us a collection of $24 \times 24 = 576$ SIRENs whose first layers are not randomly initialized, but have been selected as the best out of 100 randomly initialized layers. In the interest of training time, we reduce the number of training steps for each of the 2400 + 576 SIRENs in this experiment from 50,000 to 20,000.

\begin{figure}
    \centering
    \includegraphics[width=1.0\textwidth]{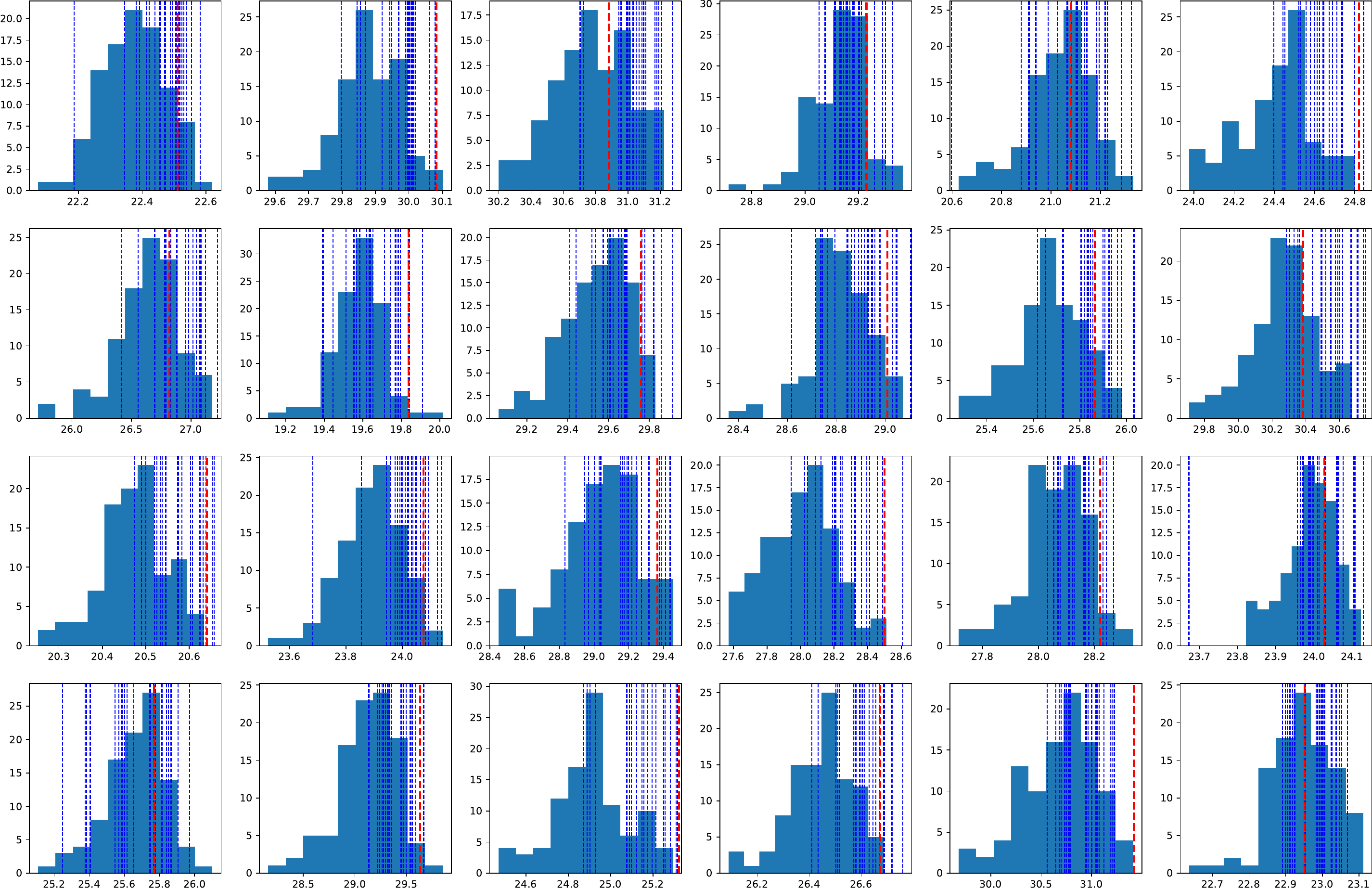}
    \caption{Histograms of the PSNR which the default COIN network reaches on each of the 24 images in the Kodak dataset across 100 different random seeds. For each image, we pick the randomized run with the best PSNR, extract its first layer, and reuse that first layer in an otherwise newly randomized network on each image. Each vertical line represents the PSNR reached by one of these retrained networks that used the best first layer out of 100. The blue vertical lines indicate that the first layer was selected from a network trained on a different image, the red vertical lines indicate the models that were retrained on the same image.}
    \label{fig:kodak-histograms}
\end{figure}

When retraining on the same image using the positional encoding with the best PSNR, our SIRENs perform \valKodakCoinImageSpecializedPEAdvantage{} PSNR better on average than with random PEs. When applying the best PE from one image to another image, average improvement in PSNR is \valKodakCoinTransferredPEAdvantage{} PSNR. This suggests that good positional encodings are somewhat transferable from image to image. Figure \ref{fig:kodak-histograms} shows the results of our experiment in detail.

\newpage
\subsubsection{Derivation of Equation \ref{eq:RMSE}}
\label{sec:RMSE}

We would like to estimate the irreducible error of our encoding error prediction problem. If we are trying to predict a variable Y from a variable X, the minimum possible RMSE that any regression model can reach is:

\[
RMSE = \sqrt{E[Var[Y | X]]}
\]

In our setting, Y is the encoding error of a SIREN as measured in PSNR, and X represents the SIREN's training inputs: width, depth, $\omega_0$ and the target image. $Y|X=x$ is the distribution of encoding errors a given SIREN setup $x$ will reach across different random seeds.

To estimate $E[Var[Y | X]]$, we randomly sample $N$ SIREN training inputs $x_1,\dots,x_N \sim X$ from our dataset, and estimate the variance $S_i^2 \approx Var[Y | X=x_i]$ for each. This gives us the following irreducible error formula:

\begin{equation}
    RMSE \approx \sqrt{\frac{1}{N} \sum_{i=1}^N S_i^2}
    \label{eq:RMSE-S2}
\end{equation}

We can estimate the variance $S^2$ of a population from $m$ samples using the unbiased sample variance estimator:

\begin{equation}
    S^2 = \frac{1}{m-1} \sum_{i=1}^m (E[y] - y_i)^2
    \label{eq:unbiased-variance}
\end{equation}

To estimate $Var[Y | X=x_i]$, we draw two samples $y_{i,1}, y_{i,2} \sim p(y|x_i)$ . So $m=2$, and $E[y] = \frac{y_{i,1} + y_{i,2}}{2}$. By substituting and simplifying, Equation \ref{eq:unbiased-variance} becomes:

\begin{equation}
    S_i^2 = \frac{1}{2} (y_{i,1} - y_{i,2})^2
    \label{eq:two-sample-unbiased-variance}
\end{equation}

Finally, we can substitute Equation \ref{eq:two-sample-unbiased-variance} into Equation \ref{eq:RMSE-S2} to get Equation \ref{eq:RMSE}.

\newpage

\newpage
\subsection{SIRENs compress the Global Color Space}
\label{sec:color-space}

\begin{figure}[t]
    \centering
    \begin{minipage}{0.25\textwidth}
        \centering
        \caption*{Ground Truth}
        \vspace{-1em}
        \includegraphics[width=\textwidth]{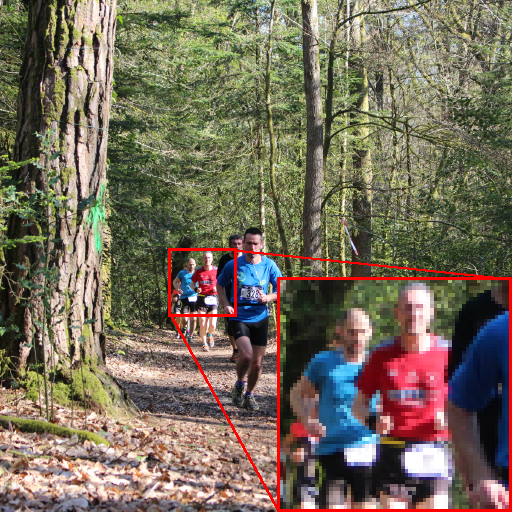}
        \includegraphics[width=\textwidth]{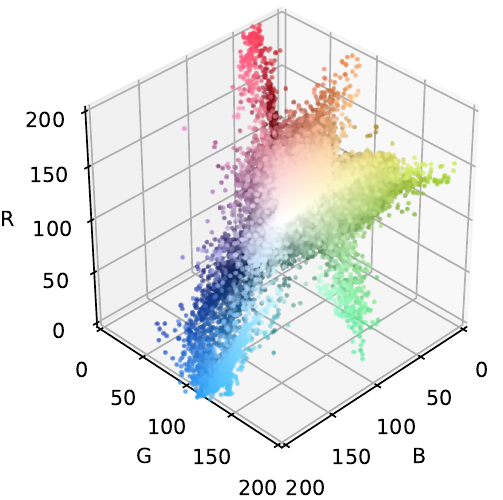}
    \end{minipage}\hspace{1px}
    \begin{minipage}{0.25\textwidth}
        \centering
        \caption*{JPEG2000 - PSNR 22.18}
        \vspace{-1em}
        \includegraphics[width=\textwidth]{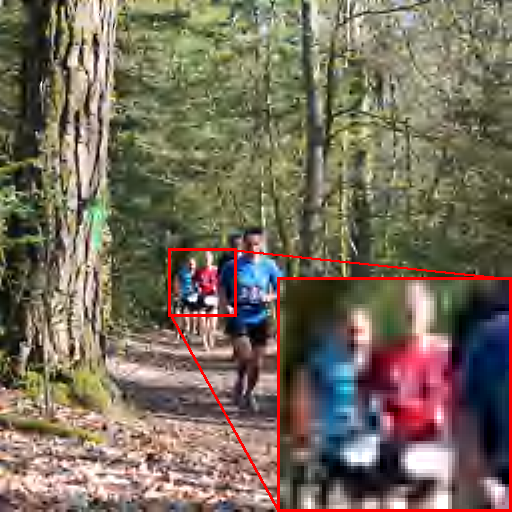}
        \includegraphics[width=\textwidth]{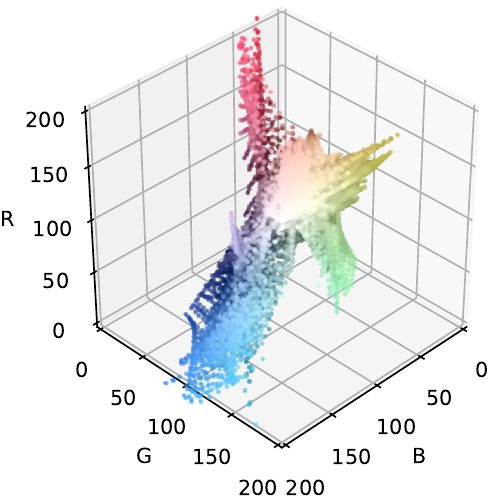}
    \end{minipage}\hspace{1px}
    \begin{minipage}{0.25\textwidth}
        \centering
        \caption*{COIN - PSNR 22.74}
        \vspace{-1em}
        \includegraphics[width=\textwidth]{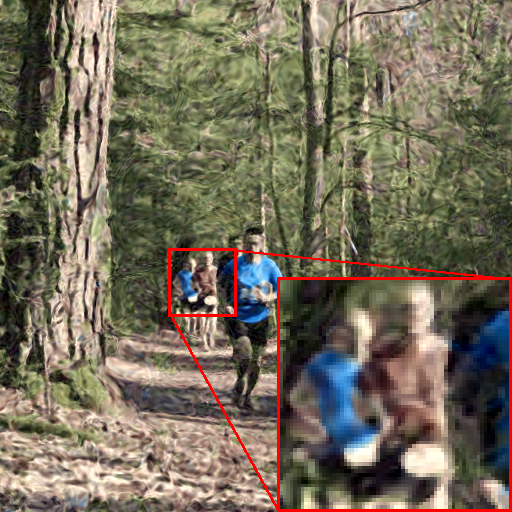}
        \includegraphics[width=\textwidth]{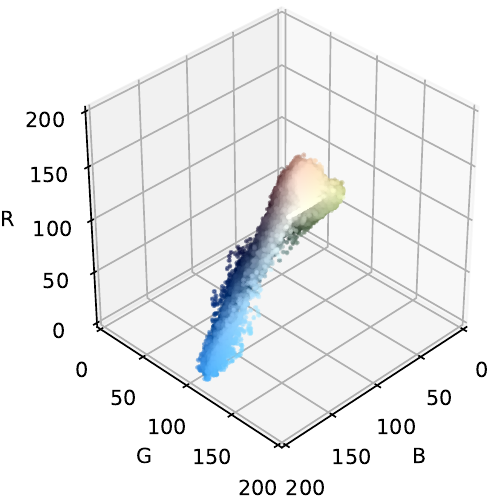}
    \end{minipage}\hspace{1px}%
    \caption{Images and their pixel-color scatter plots in RGB space. Note how the SIREN sacrifices color fidelity for spatial fidelity to achieve a higher net PSNR than JPEG2000 at 0.3 bpp. }
    \label{fig:color_scatter_plots}
\end{figure}

Figure \ref{fig:color_scatter_plots} illustrates the significant qualitative differences in the information lost by JPEG and COIN compression. At the same compression ratio, COIN better maintains the sharpness of high-contrast edges. Artifacts and distortions in COINs appear more structurally complex. Most interestingly, COINS appear to globally compress an image's color space; reducing variation in color to reduce the amount of information per pixel. We posit that this global color compression accounts for much of COIN's advantage over JPEG at low bitrates. 

If our hypothesis is correct, then our JPEG proxy models for SIREN encoding error (from Figures \ref{fig:coin-vs-jpeg-kodak} and \ref{fig:coin-vs-jpeg-mscoco}) could be further improved by accounting for the discrepancy in how COINs and JPEGs compress color space. But this remains a topic for future work.

\newpage
\begin{figure}
  \centering
  \includegraphics[width=0.6\textwidth]{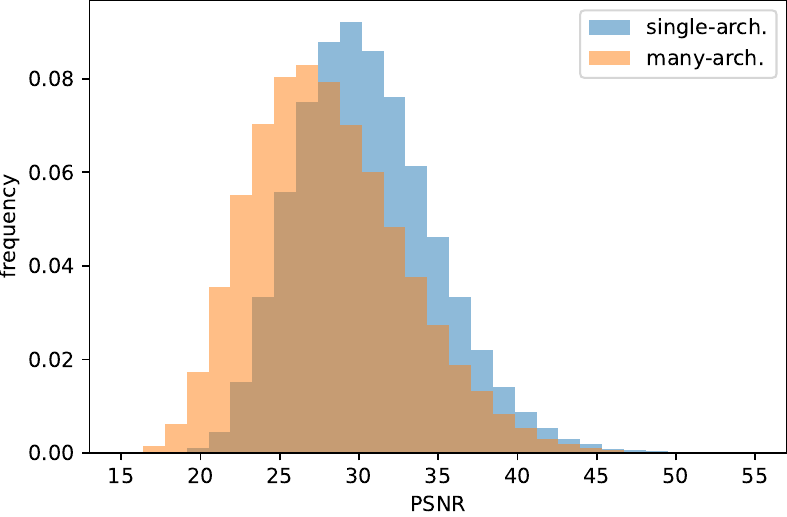}
  \caption{PSNR distribution for the single- and many-architecture SIREN datasets.}
  \label{fig:psnr_distributions}
\end{figure}

\end{document}